\documentclass[sigconf]{acmart}

\pdfoutput=1 
\usepackage{booktabs} 
\usepackage{multirow} 
\usepackage{enumitem}
\usepackage{makecell}
\usepackage[utf8]{inputenc}

\copyrightyear{2020} 
\acmYear{2020} 
\setcopyright{acmcopyright}

\acmConference[SAC '20]{The 35th ACM/SIGAPP Symposium on Applied Computing}{March 30-April 3, 2020}{Brno, Czech Republic}
\acmBooktitle{The 35th ACM/SIGAPP Symposium on Applied Computing (SAC '20), March 30-April 3, 2020, Brno, Czech Republic}
\acmPrice{15.00}

\acmDOI{10.1145/3341105.3373883}

\acmISBN{978-1-4503-6866-7/20/03}

\begin{document}

\title[]{
Better Together - An Ensemble Learner for Combining the Results of Ready-made Entity Linking Systems
}

\author{Renato Stoffalette Jo\~{a}o}
\orcid{0000-0003-4929-4524}
\affiliation{
 \institution{L3S Research Center, Leibniz University of Hannover}
  \streetaddress{Appelstra{\ss}e 9A}
  \city{Hannover}
  \state{Germany}
}
\email{joao@L3S.de}

\author{Pavlos Fafalios}
\orcid{0000-0003-2788-526X}
\affiliation{
 \institution{Institute of Computer Science, FORTH-ICS}
  \streetaddress{N. Plastira 100}
  \city{Heraklion}
  \state{Greece}
}
\email{fafalios@ics.forth.gr}

\author{Stefan Dietze}
%\orcid{}
\affiliation{%
 \institution{GESIS - Leibniz Institute for the Social Sciences}
  \city{Cologne}
  \state{Germany}
}
\additionalaffiliation{
 \institution{L3S Research Center, Leibniz University of Hannover}
  \streetaddress{Appelstra{\ss}e 9A}
  \city{Hannover}
  \state{Germany}
}
\email{stefan.dietze@GESIS.org}

\renewcommand{\shortauthors}{R. Stoffalette Jo\~{a}o et al.}

\begin{abstract}
Entity linking (EL) is the task of automatically identifying entity mentions in text and resolving them to a corresponding entity in a reference knowledge base like Wikipedia. Throughout the past decade, a plethora of EL systems and pipelines have become available, where performance of individual systems varies heavily across corpora, languages or domains. 
Linking performance varies even between different mentions in the same text corpus, where, for instance, some EL approaches are better able to deal with short surface forms while others may perform better when more context information is available.
To this end, we argue that performance may be optimised by exploiting results from distinct EL systems on the same corpus, thereby leveraging their individual strengths on a per-mention basis. 
In this paper, we introduce a supervised approach which exploits the output of multiple ready-made EL systems by predicting the correct link on a per-mention basis. Experimental results obtained on existing ground truth datasets and exploiting three state-of-the-art EL systems show the effectiveness of our approach and its capacity to significantly outperform the individual EL systems as well as a set of baseline methods.

\end{abstract}

% The code below should be generated by the tool at: http://dl.acm.org/ccs.cfm



\maketitle

\section{Introduction}
\label{sec:introduction}

Entity linking (EL), or named entity recognition and disambiguation (NERD), is the task of determining the identity of entity mentions in text, thus linking a mention to an entity in a reference Knowledge Base (KB) like Wikipedia \cite{shen2015entity}. 
For example, in the sentence \lq\lq{}{}Jordan played for the Wizards\rq\rq{}{}, a typical EL system would link the term \lq\lq{}{}Jordan\rq\rq{}{} to the Wikipedia page of the basketball player \textit{Michael Jordan} and the term \lq\lq{}{}Wizards\rq\rq{}{} to the Wikipedia page of the USA basketball team \textit{Washington Wizards}.

EL is a crucial task of relevance for a wide variety of applications, such as information retrieval \cite{raviv2016document}, document classification \cite{ni2016semantic},
or topic modelling \cite{chen2016probabilistic}.
Usually, high precision and recall are required if EL results are to have a positive impact on any such application.

However, EL remains a challenging task. EL systems differ along multiple dimensions and are evaluated over different datasets \cite{shen2015entity}, while their performance varies significantly across domains and corpora \cite{roder2017gerbil}. For instance, evaluations using the GERBIL benchmark \cite{roder2017gerbil} have shown that the performance of EL systems is highly affected by the characteristics of the datasets, such as the number of entities per document, the average document length, or the salient entity types \cite{usbeck2015evaluating}. 
Thus, general-purpose EL remains a challenging task, where no single system has yet emerged as de-facto-standard across corpora and EL scenarios.

EL performance also varies strongly on each individual mention in the same corpus. As we will show in our evaluation (Table \ref{tbl:ELperformance}), the F1 score of three established  EL systems (TagMe, Ambiverse, Babelfy) on the popular CONLL dataset \cite{hoffart2011robust} ranges between 63.5\% - 74.3\% with an upper bound performance of 90.6\% when selecting the most correct outputs of all three systems. 
This underlines that selecting the EL system on a per mention-basis rather than for a particular corpus, can significantly increase the EL performance. 
However, the selection of the most suitable system for a given mention remains a challenge.
Prior works have shown that mentions which are difficult to link often share common characteristics \cite{fafalios2019SameButDifferent,hoffart2012kore}, which include ambiguity, indicated by a large number of candidates, mentions of long-tail entities which are not well represented in reference KBs, or mentions recognised in short documents with very limited context information. 

Drawing on these observations, we argue that effective features can be derived from the corpus, the mention or the surface form to be linked, in order to predict the best-performing EL system on a per-mention-basis using supervised models. 
In this work we introduce an ensemble learning approach towards exploiting the EL capabilities of a set of ready-made EL systems not only for improving recall, but also to improve precision by predicting the most correct EL system considering the particular characteristics of each particular mention.
We focus on exploiting ready-made (end-to-end) EL systems that are used as {\em black-box} systems using their default (suggested) configuration and without any corpus-specific training or tuning, because such systems are widely used in different contexts by also non-expert users.

We apply our approach to three established EL datasets and demonstrate significant performance improvements compared to both the individual EL systems and six baseline strategies. 
Specifically, when considering the largest dataset (CONLL), our ensemble-based method significantly outperforms the best performing individual EL system by 10\% of F1 score, as well as the top performing baseline by 5\%. 

In a nutshell, we make the following contributions: 

\begin{itemize}
    \item We introduce the problem and a novel approach towards \textit{Meta Entity Linking}, in short \textit{MetaEL}, where outputs of multiple end-to-end EL systems are combined using an ensemble learning method for providing an improved set of entity links for a given corpus. 
    \item  We propose a diverse set of features which give suitable signals for predicting the EL system that can provide the correct link for a given mention, and build supervised classifiers which are used as part of an automated \textit{MetaEL} pipeline. 
    \item Using existing ground truth datasets and a set of three established and ready-made EL systems, we first provide detailed annotation and agreement statistics which demonstrate the potential performance improvement that an effective \textit{MetaEL} method can provide.  
    Then, we report the resulting EL performance gain of the proposed supervised approach as well as evaluation results on the prediction task per se and the importance of the devised features, discussing also the limitations of our approach.
\end{itemize}

The rest of the paper is organised as follows: 
Section \ref{sec:problem} formulates the \textit{MetaEL} problem and provides an overview of our approach.
Section \ref{sec:classification} details how supervised classification can be used for the problem at hand. 
Section \ref{sec:evalSetup} describes the evaluation setup. 
Section \ref{sec:evalResults} reports the evaluation results.
Section \ref{sec:rw} discusses related works and the difference of our approach. 
Finally, Section \ref{sec:conclusion} concludes the paper and discusses interesting directions for future research.

\section{Problem Definition}
\label{sec:problem}

We first define the notion of \textit{entity linking}: 

\begin{definition}[Entity Linking]
Given a corpus of documents $D$, e.g., a set of news articles,
and a KB $K$, e.g., Wikipedia, that describes information for a set of entities $E$ (like persons, locations, organisations, etc.), the task of \textit{Entity Linking (EL)} aims at providing a set of \textit{entity annotations} $A$ of the form  $\langle m, e\rangle$, where $m$ is an entity mention recognised in a document of $D$ and $e$ is an entity in $E$ that determines the identity of $m$. An entity mention $m$ is of the form $\langle d, s, p\rangle$ where $d$ is a document in $D$, $s$ is a surface form in $d$ (a word or a sequence of words representing an entity), and $p$ is the position of $s$ in $d$. Thus, an entity mention is always identified by a document, a surface form, and a position. We denote by $M$ the set of all recognised entity mentions $\langle d, s, p\rangle$.
\end{definition}

Our approach exploits a set of $n$ EL systems $(l_1, .. , l_n)$ which operate on the same reference KB $K$ and are applied to the same corpus $D$.
The output is $n$ sets of entity annotations $\mathcal{A} = (A_1, \dots, A_n)$, corresponding to $n$ sets of entity mentions $\mathcal{M} = (M_1, \dots, M_n)$, each one produced by a different EL system $l_i$. The size of each set of entity mentions $M_i$ can be different, since each system might have recognised different entity mentions. 

\begin{definition}[Meta Entity Linking]
Assuming that, for a given corpus $D$, we have $n$ sets of entity mentions $\mathcal{M} = (M_1, \dots, M_n)$ and $n$ sets of corresponding entity annotations $\mathcal{A} = (A_1, \dots, A_n)$, produced by $n$ different EL systems $(l_1, .. , l_n)$, the task of \textit{Meta Entity Linking}, for short \textit{MetaEL}, aims at providing a unified set of entity annotations $A_u$, where for each entity mention $m \in (M_1 \cup ... \cup M_n)$, the most correct annotation is selected from $(A_1 \cup ... \cup A_n)$.
\end{definition}

The solution proposed in this paper, called {\em MetaEL+}, is based on {\em supervised classification}. We propose two variations, one focusing on high recall (\textsc{LOOSE}) and a more selective one focusing on high precision (\textsc{STRICT}). In both approaches, if at least two of the considered EL systems have recognised and disambiguated the same entity mention $m$, we predict the system to consider using a \textit{multi-label classifier}. 
If only one EL system has recognised an entity mention, the \textsc{STRICT} approach predicts if the provided entity link is correct using a system-specific \textit{binary classifier}. On the contrary, the \textsc{LOOSE} approach includes all annotations recognised by only one of the systems, thus focusing on high recall.

\section{MetaEL+}
\label{sec:classification}

This section describes the \textit{features}, \textit{classifiers} and \textit{labelling} methods used by the proposed \textit{MetaEL+} approach.  

\subsection{Features}
\label{subsec:features}

We propose a set of features that can be easily computed for arbitrary corpora, i.e., we are not interested in features that, for example, require special metadata information about the documents. Inspired by related works on EL which study different factors that affect the performance of EL systems \cite{shen2015entity,roder2017gerbil}, as well as by the observed characteristics of mentions that fail to be disambiguated correctly, we consider features of the following categories: 
i) \textit{surface form-based} (features related to the word or sequence of words representing an entity), 
ii) \textit{mention-based} (features related to the mention recognised in a document, in a specific position), and 
iii) \textit{document-based} (features related to the document containing the mention).
Below we detail each one of them.

\vspace{1.5mm} \noindent 
{\em \large Surface Form-based Features \normalsize}
\vspace{0.5mm} \noindent

\noindent
\textbf{Number of words ($s_{words}$):} the number of surface form's words. Our intuition is that an EL system may perform better/worse on unigram surface forms that are usually more ambiguous than surface forms with more than one word. 

\noindent
\textbf{Frequency ($s_{f}$):} the number of surface form's occurrences within the document. Our intuition is that more occurrences implies that the document topic is closely related to the surface form, indicating more representative context to facilitate its disambiguation. 

\noindent
\textbf{Document frequency ($s_{df}$):} the number of documents in the corpus $D$ containing at least one occurrence of the surface form. Higher value implies popularity of the surface form, suggesting that more context is available about it which can facilitate its disambiguation by EL systems. 

\noindent
\textbf{Number of candidate entities ($s_{cand}$):} the number of candidate entities in the reference KB (obtained by exploiting Wikipedia hyperlinks with anchor texts pointing to entities). We anticipate that individual EL systems may perform better/worse on ambiguous mentions having a high number of candidate entities. 

\noindent
\textbf{Surface form's correct disambiguations per EL system ($s_{corr}$):} number of times the surface form has been disambiguated correctly by a specific EL system on the given training dataset.
Our intuition is that an EL system which has disambiguated correctly a particular occurrence of a surface form is more likely to disambiguate correctly a different occurrence of the same term.

\noindent
\textbf{Surface form's ratio of correct disambiguations per EL system ($s_{ratio}$):}  ratio of times the surface form has been disambiguated correctly by a specific EL system on the given training dataset. It is computed as the number of correct disambiguations divided by the sum of correct and wrong disambiguations. 
Similarly to the previous feature, our intuition is that an EL system which performed well on a number of occurrences of a particular surface form will perform well on the same term in the future.

\vspace{1.5mm} \noindent 
{\em \large Mention-based Features \normalsize}
\vspace{0.5mm} 

\noindent
\textbf{Mention's normalised position ($m_{pos}$):} the mention's normalised position in the document, computed as the number of characters from the start of the document divided by the total number of characters in the document. Entities appearing early in the document are usually salient and representative for the document, indicating more representative context to facilitate their disambiguation. 

\noindent
\textbf{Mention's sentence size ($m_{sent}$):} the number of characters of the sentence containing the mention, specifically the length of the text between two punctuation marks containing the mention (considering only the punctuation marks ".", "!", "?", ";"). Whereas an EL system may exploit the sentence containing the mention for disambiguating the entity, larger sentences may indicate more representative context for a particular mention. 

\vspace{1.5mm} \noindent 
{\em \large Document-based Features \normalsize}
\vspace{0.5mm} 

\noindent
\textbf{Document size ($d_{words}$):} the number of words of the document containing the mention. We anticipate that the document length may provide signals for EL system performance with some approaches being able to deal better with short documents (containing more concise but less context), while others with longer documents.

\noindent
\textbf{Document's recognised entities ($d_{ents}$):} the total number of entities recognised in the document containing the mention. Given that EL systems tend to jointly disambiguate entities, some EL systems may perform better in the presence of a larger amount of recognised entities.

\subsection{Classifiers}
\label{subsec:models}

Since more than one EL system can provide the correct entity link for a recognised entity mention, we model the problem as a \textit{multi-label classification} task \cite{tsoumakas2007multi} where multiple labels (systems) may be assigned to each instance (entity mention). 
We experimented with a large number of different methods using the MEKA framework \cite{MEKA} (an open source implementation of several methods for multi-label classification), trying also different base classifiers for each method, including Naive Bayes (NB), Logistic Regression (LR), J48, Random Forest (RF), and Sequential Minimal Optimisation (SMO). 
In our evaluation (Sect. \ref{sec:evalResults}), we report results only for the top performing method: \textit{Binary Relevance} using {\em RF} as the base classifier.

As regards the case where only one EL system has recognised an entity mention $m$, a STRICT approach (as described in Sect. \ref{sec:problem}) needs to predict if the provided entity link is correct. For this, we need $n$ binary classification models, one for each considered EL system $(l_1, .. , l_n)$, where the class label is either \textit{true} (the EL system provides the correct entity link for $m$), or \textit{false} (the EL system does not provide the correct entity link for $m$). We experimented with many different classification models, including NB, LR, J48, RF, KNN, and SMO. We report results only for SMO which consistently had the best performance across datasets. 

\subsection{Training and Labelling}
\label{subsec:training}

For training supervised classifiers on the prediction tasks, one can generate training instances using manual labelling (e.g., from domain experts) \cite{gerber2013real}, crowd-sourcing \cite{bontcheva2017crowdsourcing}, or existing ground truth datasets \cite{roder2017gerbil}. In our experiments, we make use of existing ground truth datasets (more in Sect. \ref{sec:evalSetup}). After annotating the documents of the training corpus, we compute the feature values for each mention that exists in the ground truth and assign the corresponding class labels. 
For the \textit{multi-label classifier}, we label the training instances by simply considering the systems that managed to correctly disambiguate the mention. 
For each \textit{binary classifier}, we make use of only the annotations produced by the corresponding EL tool and label the training instances as either \textit{true} or \textit{false}.

\section{Evaluation Setup}
\label{sec:evalSetup}

We evaluate the EL performance of \textit{MetaEL+} for a given set of ready-made EL tools. Since the previously introduced prediction task is an integral element of \textit{MetaEL+}, we also evaluate the prediction performance of the proposed supervised classifiers.

\subsection{Datasets}

We need datasets for which enough ground truth (GT) annotations are provided for training a supervised classifier. We considered the following three datasets, each one containing at least 1,000 training annotations: i) \textit{CONLL} (GT annotations for 1,393 Reuters articles \cite{hoffart2011robust}), ii) \textit{IITB} (GT annotations for 107 text documents drawn from popular web pages about sports, entertainment, science, technology, and health \cite{kulkarni2009collective}), iii) \textit{NEEL} (GT annotations for >9,000 tweets, provided by the 2016 NEEL challenge \cite{Microposts2016}). 
Several other GT datasets have not been considered because of their very small size (e.g. \textit{ACE2004}, \textit{Aquaint}, \textit{KORE50}, \textit{Meij}, \textit{MSNBC}). 
CONLL and NELL are already split into training and test sets. For IITB we considered the first 90\% of the provided annotations for training and the remaining 10\% for test (thus one can reproduce the results). 
In all datasets, we do not consider GT annotations pointing to \textit{NULL} or \textit{OOKB} (out of knowledge base). Table \ref{tbl:datasetsStats} shows the number of documents and annotations per dataset used for training and test (considering only the documents having at least one GT annotation).

\begin{table}[H]
\vspace{-2mm}
\centering
\renewcommand{\arraystretch}{0.7}
\setlength{\tabcolsep}{4.5pt}
\caption{Ground truth datasets main statistics.}
\vspace{-4mm}
\label{tbl:datasetsStats}
\small
\begin{tabular}{rrrrr}	
     \toprule
     Dataset     &  \makecell{\#Train docs} &  \makecell{\#Train annots} & \makecell{\#Test docs}  & \makecell{\#Test annots}\\  
     \midrule
     CONLL      & 1,162   & 23,332   & 231    & 4,485 \\ 
     IITB       & 90      & 10,847   & 13     & 1,174 \\ 
	 NEEL       & 3,342   & 6,374    & 291    & 736   \\
	 \bottomrule
\end{tabular}
\vspace{-2.5mm}
\end{table}

\subsection{Entity Linking Tools}
\label{subsec:tools}

We deployed three popular state-of-the-art EL tools:
\textit{Ambiverse} (previously AIDA) \cite{hoffart2011robust},
\textit{Babelfy} \cite{moro2014entity}, and \textit{TagMe} \cite{ferragina2010tagme}. 
These tools were selected because: i) they are end-to-end (ready-made) tools that can be easily used out-of-the-box, and ii) they are accessible through public APIs, thus one can directly use them. Moreover, they have been widely used in different contexts (each one having >400 citations). Other EL systems, including more recent ones that make use of neural models, have not been considered because they do not satisfy these criteria.
For Ambiverse, we used its default configuration. For Babelfy, we used the configuration suggested by the Babelfy developers.\footnote{The configuration is available at: \url{https://goo.gl/NHXVVQ}} 
For TagMe we used its default configuration and a confidence threshold of 0.2 to filter out low quality annotations.

\begin{comment}
\begin{table}[!h]
%\vspace{-1.5mm}
\centering
\renewcommand{\arraystretch}{0.7}
\setlength{\tabcolsep}{4.5pt}
\caption{Performance of the used EL tools on CONLL \cite{hoffart2011robust}.}
\vspace{-3mm}
\label{tbl:perfEntLink}
\small
\begin{tabular}{p{1.6cm} p{1.6cm} p{1.2cm} p{0.8cm}}		
     \toprule
     \textbf{System}     & \textbf{Precision (\%)} & \textbf{Recall (\%)}   & \textbf{F1 (\%)}    \\ 
     \midrule
     Ambiverse  & 80.7    & 64.7     & 71.8  \\ 
	 Babelfy    & 81.5    & 68.2     & 74.3  \\ 
	 TagMe      & 78.7    & 53.2     & 63.5   \\
	 \bottomrule
\end{tabular}
\vspace{-3mm}
\end{table}
\end{comment}

\subsection{Baseline and MetaEL+ Methods}
\label{subsec:baselines}

Since the objective of MetaEL is the selection of output from multiple EL tools for achieving a better performance, each of the used tools (Ambiverse, Babelfy and TagMe) is considered a different and naive baseline. In addition, considering the agreement of the tools on the provided entity (majority vote) or their overall performance in a ground truth dataset, are two other predictive baselines \cite{corcoglioniti2016microneel,si2005boosting}. As regards the \textit{MetaEL} problem per se, \cite{ruiz2015combining} proposes a weighted voting scheme which ranks the candidate entities by considering the performance of the tools on a so-called ranking corpus (CONLL).
The considered baselines are summarised below: 

\begin{itemize}[noitemsep,nolistsep]
    \item Each considered EL system (\textbf{Ambiverse}, \textbf{Babelfy}, \textbf{TagMe}).
    \item \textbf{Random}: select one of the tools randomly.
    \item \textbf{Best System}: select the link provided by the system with the highest overall performance in the ground truth dataset.
    \item \textbf{Majority+Random}: select the link provided by the majority of the tools. If all tools provide a different link, a random one is selected. 
    \item \textbf{Majority+Best}: select the link provided by the majority of the tools. If all tools provide a different link, the system with the highest overall performance is selected. This method is similar to the \textit{rule-based} method of \cite{corcoglioniti2016microneel}.
    \item \textbf{Weighted Voting}: the annotations are combined through the weighted voting scheme described in \cite{ruiz2015combining}. If the score is lower than the maximum precision for all annotators on the ranking corpus, the annotation is not considered. 
    \item \textbf{Weighted Voting All}: the annotations are combined through the weighted voting scheme described in \cite{ruiz2015combining}, however without filtering out annotations with a score lower than the maximum precision for all annotators.
\end{itemize}

We compare the performance of the above mentioned baselines with the following two \textit{MetaEL+} approaches: 

\begin{itemize}[noitemsep,nolistsep]
    \item \textbf{MetaEL+$_{\text{LOOSE}}$}: a multi-label binary relevance classifier (with RF as the base classifier) is used when more than one tool provide a link for the same mention. We use the implementation and default configuration of MEKA 1.9.3 \cite{MEKA}.
    When more than one system is predicted, we consider the \textit{prediction confidence} scores provided by the classifier for each class. 
    In case of equal scores, we select the system with the highest overall performance in the training dataset. 
    If only one EL system has recognised a mention, we trust it and assign the entity provided by this system. 
    
    \item \textbf{MetaEL+$_{\text{STRICT}}$}: the same multi-label classifier from the MetaEL+$_{\text{LOOSE}}$ approach is used for cases where more than one tool provide a link for the same mention. However, this method is more selective: when a mention is recognised by only one EL system, a system-specific SMO binary classifier is used for predicting if the provided entity link is correct. 
\end{itemize}

\subsection{Evaluation metrics}

\subsubsection{Evaluating EL performance}
We make use of the following metrics: \textbf{Precision (P)} (number of correctly disambiguated mentions divided by the number of recognised mentions), \textbf{Recall (R)} (number of correctly disambiguated mentions divided by the total number of not null annotations in the ground truth), and \textbf{F1 score (F1)} (harmonic mean of precision and recall).

\subsubsection{Evaluating the classification performance}
\label{subsubsec:evalMetricsClassPerf}

Evaluation metrics for multi-label classification are inherently different from those used in single-label classification (like binary or multi-class) \cite{tsoumakas2007multi}. 
We report results for the following metrics: \textbf{Jaccard Index} (number of correctly predicted labels divided by the union of predicted and true labels), \textbf{Hamming Loss} (fraction of the wrong labels to the total number of labels), \textbf{Exact Match} (percentage of samples that have all their labels classified correctly), \textbf{Per-class Precision} (P), \textbf{Recall} (R) and \textbf{F1 score} (if $TL$ denotes the true set of labels for a given class and $PL$ the predicted set of labels for the same class, then $P = \frac{TL \cap PL}{PL}, R = \frac{TL \cap PL}{TL}$, and $F1 = \frac{ 2\cdot P \cdot R}{P+R}$).

Since, in our problem, the prediction of any of the tools that provides the correct entity is adequate, we also report the accuracy of the classifiers in each dataset when considering if the correct entity is provided by the predicted system. Based on this, we define \textbf{Real Prediction Accuracy} as the number of predictions for which the predicted system provides the correct entity divided by the total number of predictions. 

Finally, for measuring the performance of the three binary classifiers used by the \textsc{STRICT} approach, we consider the \textbf{per-class P}, \textbf{R} and \textbf{F1 score} as well as the \textbf{macro-averaged F1 score}.

\section{Evaluation Results}
\label{sec:evalResults}

\subsection{Annotation Statistics and Upper Bound Performance}

\subsubsection{Annotation and agreement statistics}
Table \ref{tbl:annotStats} provides detailed statistics about the annotations of the test datasets using the three EL tools. These statistics can help us better understand the characteristics of the datasets and the behaviour of the considered tools.

\begin{table}[!h]
%\vspace{-2mm}
\centering
\renewcommand{\arraystretch}{0.7}
\setlength{\tabcolsep}{1.0pt}
\caption{Annotation statistics of the test datasets.}
\vspace{-4mm}
\label{tbl:annotStats}
\small
\begin{tabular}{p{4.3cm}rrrr}	
\toprule
     & \textbf{CONLL}  & \textbf{IITB}   & \textbf{NEEL}  \\   
\midrule
Total number of GT annotations:                                & 4,485   & 1,174     & 736   \\   
Ambiverse annotations:                                         & 4,169   & 390       & 66   \\   
Babelfy annotations:                                           & 9,578   & 868       & 246    \\  
TagMe annotations:                                             & 4,626   & 355       & 801   \\  
\midrule
GT mentions recognised by 0/3 tools:                           & 295 (6.6\%)     & 694  (59.1\%)   & 456 (62.0\%)  \\ 
 GT mentions recognised by 1/3 tools:                           & 468 (10.4\%)    & 227 (19.3\%)     & 225 (30.6\%) \\ 
\hspace{1.0mm} \textit{Correct entity is provided:}          & 337 (72\%)     & 63 (27.8\%)      & 141 (62.7\%)   \\ 
GT mentions recognised by 2/3 tools:                           & 1,251 (27.9\%)  & 125 (10.6\%)  & 43 (5.8\%) \\ 
\hspace{1.0mm} The 2 tools provide the same entity:              & 950  (75.9\%)   & 88 (70.4\%)         & 32 (74.4\%) \\ 
\hspace{1.0mm} The 2 tools provide different entities:           & 301 (24.1\%)     & 37 (29.6\%)         & 11 (25.6\%) \\ 
\hspace{1.0mm} \textit{Correct entity is provided:}     & 1,061 (84.8\%)  & 103 (82.4\%)           & 37 (86\%)  \\
GT mentions recognised by 3/3 tools:                           & 2,471 (55.1\%)  & 128 (10.9\%) & 12 (1.6\%)   \\ 
\hspace{1.0mm} 3/3 tools provide the same entity:  & 1,786 (72.3\%)   & 82 (64.1\%)       & 8 (66.7\%)    \\
\hspace{1.0mm} 2/3 tools provide the same entity:  & 618 (25\%)     & 37 (28.9\%)       & 3 (25\%)     \\ 
\hspace{1.0mm} Each tool provides a different entity: & 67 (2.7\%)      & 9 (7\%)        & 1 (8.3\%)    \\ 
\hspace{1.0mm} \textit{Correct entity is provided:}     & 2,314 (93.6\%)   &  119 (93\%)     & 12 (100\%)\\
\bottomrule
\end{tabular}
\vspace{-3mm}
\end{table}

The first four rows show the total number of GT annotations in each dataset and the number of annotations produced by each of the considered EL tools.
The next rows show the number of GT mentions recognised by zero, only one, two, or all three tools, as well as the number of mentions for which at least one of the tools provides the correct entity and the agreement of the tools in the provided entities. 
We notice different patterns across the datasets. Concerning CONLL, for instance, we notice that the majority of GT mentions were recognised by all three tools (55.1\%), followed by mentions recognised by 2/3 tools (27.9\%). 
On the contrary, for IITB and NEEL, we see that the majority of GT mentions were not recognised by at least one system (59\% and 62\%, respectively). 
Based on these numbers, we expect a high improvement of recall when we combine the annotations of the three tools since a much larger number of GT annotations are expected to have been recognised by at least one system. 
Moreover, we see that for a quite high percentage of mentions recognised by only one EL tool, the provided entity is not correct (28\% in CONLL, 72\% in IITB, 37\% in NEEL). Thus, an effective \textit{MetaEL} method focusing on high precision should avoid including these annotations in the unified set of entity annotations.

With respect to the agreement of the tools on the provided entities, we noticed that when more than one system provides an entity for the same mention, the tools usually agree on the entity.
Nevertheless, we notice that for CONLL there is a high percentage of GT mentions where the tools disagree and provide different entities (22\% of all GT annotations). This percentage is 7\% for IITB and around 2\% for NEEL. The problem with these two datasets (especially with NEEL) is that the percentage of GT mentions recognised by zero or only one system is very high (78\% for IITB and 93\% for NEEL), in contrast to CONLL where the percentage is only 17\%. 
If we consider only the GT mentions for which at least two tools provide a link, then the percentage of mentions that need prediction is again high (26.5\% for CONLL, 32\% for IITB, and 27.3\% for NEEL).
For all these mentions, an effective \textit{MetaEL} method needs to predict the system that can provide the correct entity link. 

The above analysis shows that there is (i) a high percentage of mentions for which the EL tools provide different entities, and (ii) a high percentage of mentions for which no EL tool provides the correct entity. This means that predicting the system to consider or the correctness of an annotation can significantly improve the overall EL performance.

\subsubsection{Upper bound performance}
\label{sbusubsec:ubp}

Given the GT of each dataset, we can compute the performance of an ideal \textit{MetaEL} system that always makes a correct prediction (and thus no other method can provide better results).
The first row in Table  \ref{tbl:ELperformance}
shows the upper bound performance for each of the considered datasets. Comparing the upper bound performance for CONLL with the performance of the three individual tools on the same dataset (rows 2-4 in Table \ref{tbl:ELperformance}), we notice that \textit{MetaEL} can highly increase the F1 score from 74.3\% (of Babelfy, the top performing system) to 90.6\%, i.e., >15 percentage points (or 22\% increment).
With respect to the other datasets, we see that the F1 score of the upper bound performance is relatively low. As we will see below, the reason is the low recall achieved by all tools. Nevertheless, the F1 score of the upper bound performance is much higher than that of the top performing individual system in each case (33\% increment for IITB and 18\% for NEEL). 
These results provide a good motivation for an effective \textit{MetaEL} method that can achieve a high performance as close to the upper bound performance as possible. 

\vspace{-2mm}
\subsection{Entity Linking Performance}
\label{subsec:metaelPerf}
Table \ref{tbl:ELperformance} shows the EL performance of all approaches on the different datasets. The first row shows the upper bound performance and the next three rows the performance of the individual EL tools. The next six rows show the performance of the six baseline methods and the last two rows the performance of our two \textit{MetaEL+} methods.

\begin{table*}[] 
\vspace{-3mm}
\centering
\caption{Entity linking performance.}
\vspace{-4mm}
\renewcommand{\arraystretch}{0.7}
\setlength{\tabcolsep}{4.5pt}
\small
\label{tbl:ELperformance}
\begin{tabular}{@{}lccc|ccc|ccc@{}} \hline
    \multirow{2}{*}{\textsc{\textbf{Method}}} 
    & \multicolumn{3}{c}{\textsc{\textbf{CONLL-Test}}} 
    & \multicolumn{3}{c}{\textsc{\textbf{IITB-Test}}} 
    & \multicolumn{3}{c}{\textsc{\textbf{NEEL-Test}}} \\
 \multicolumn{1}{c}{}  & P (\%) & R (\%) & F1 (\%) & P (\%) & R (\%) & F1 (\%) & P (\%) & R (\%) & F1 (\%) \\ \hline 
 
 \textsc{Upper bound} & 
 100.0 & 82.8 & 90.6 &   
 100.0 & 24.3 & 39.1 & 
 100.0 & 25.7 & 40.9 \\ \hline \hline

  \textsc{Ambiverse} &   
  80.7 & 64.7 & 71.8 &
  \textbf{85.2} & 17.7 & 29.3 &   
  \textbf{76.6} & 4.9 & 9.2 \\

  \textsc{Babelfy} &    
  81.5 & 68.2 & 74.3 &       
  42.7 & 13.7 & 20.8 &    
  64.4 & 3.9 & 7.4 \\
 
 \textsc{TagMe} &   
 78.7 & 53.2 & 63.5 &         
 72.3 & 14.9 & 24.7 &   
 67.5 & 23.4 & 34.7 \\ \hline

  \textsc{Random} &   
  79.3 & 74.1 & 76.7 &     
  52.7 & 21.6 & 30.6 &    
  64.3 & 24.5 & 35.4 \\

  \textsc{Best System} 	 &   
  80.3 & 75.0 & 77.5 &    
  57.9 & \textbf{23.7}& 33.6 &   
  65.7 & \textbf{25.0} & \textbf{36.2} \\ 

 \textsc{Majority+Random}	 &   
 80.8 & 75.5 & 78.0 &    
 54.7 & 22.4 & 31.8 &    
 65.4 & 24.9 & 36.0 \\

 \textsc{Majority+Best}  	 &   
 80.5 & 75.3 & 77.8 &    
 57.7 & 23.6 & 33.5 &    
 65.7 & \textbf{25.0} & \textbf{36.2} \\

  \textsc{Weighted Voting}  	 &  
  80.8 & 72.5  & 76.4 &    
  44.8 & 17.3  & 25.0 &   
  63.5 & 22.7  & 33.4 \\

  \textsc{Weighted Voting All}  	 &  
  80.3 & 75.0  & 77.5 &   
  48.0 & 19.4  & 27.6  &   
  65.7 & \textbf{25.0}  & \textbf{36.2} \\ \hline

  \textsc{MetaEL+$_{\text{LOOSE}}$} 	 &
   84.8 & \textbf{79.2} & \textbf{81.9} & 
   57.7 & 23.6 & 33.5 & 
   65.7 & \textbf{25.0} & \textbf{36.2} \\
 
  \textsc{MetaEL+$_{\text{STRICT}}$} 	 &
  \textbf{86.6} &  75.2 & 80.5 &
  84.8 &  22.3 & \textbf{35.3} &
  73.0 &  9.9  & 17.5  \\
  
  \hline
\end{tabular}
\vspace{-3mm}
\end{table*}

To calculate the statistical significance of our results, we divided the test set of each dataset into 20 disjoint splits of equal number of annotations, and computed the F1 score on each split for each method (similar to the approach in \cite{fang2014entity}). Two-tail paired t-test was then applied to determine if the F1 scores of our methods and the baselines are significantly different.

First, we notice that the performance of the individual EL tools varies across datasets. As regards CONLL, Babelfy is the top performing tool and TagMe the tool with the worst performance (in terms of F1 score). For IITB, Ambiverse is the top performing tool and Babelfy the worst one. For NEEL, TagMe is the tool with the best performance and Babelfy the one with the worst performance. 
These results validate our motivation that the performance of EL systems varies across datasets.

Regarding the performance of the proposed MetaEL+ approaches, we notice that our \textsc{LOOSE} approach achieves the highest F1 score in CONLL (the largest and most reliable dataset), outperforming the top performing individual system by 10\% (from 74.3\% to 81.9\%) and the top performing baseline by 5\% (from 78\% to 81.9\%). In more detail, recall of the top performing EL system (\textsc{Babelfy}) is improved from 68.2\% to 79.2\% (very close to the upper bound performance) and at the same time precision is improved from 81.5\% to 84.8\%. This is very promising given that, usually, improvement in recall affects precision negatively. 
With respect to the baseline methods, recall of the top performing baseline (\textsc{Majority+Random}) is improved from 75.5\% to 79.2\% and precision from 80.8\% to 84.8\%. All these improvements are statistically significant for $\alpha$-level = 0.05.
We also see that, with a drop of recall to 75.2\%, precision can be further improved to 86.6\% using the \textsc{STRICT} approach. Here we would expect a higher improvement of precision, which means that the binary classifiers are not probably very effective in distinguishing \textit{true} from \textit{false} instances (this hypothesis is validated below). % in Sect. \ref{subsec:predPerformance}).

In IITB, our \textsc{MetaEL+$_{\text{STRICT}}$} approach achieves the highest F1 score, outperforming the top performing EL tool (\textsc{Ambiverse}) by 20.5\% and the top performing baseline by 5\%. % (from 29.3\% to 35.3\%)(from 33.6\% to 35.3\%). 
Our method combines a high recall (compared to that of the individual systems) with a very high precision (84.8\%). Precision, in particular, is improved compared to the best baseline (\textsc{Best System}) by 46.5\% while recall slightly drops from 23.7\% to 22.3\%. 

Finally, in NELL we notice that our \textsc{LOOSE} approach and four of the baseline systems achieve the same performance. This is not surprising given the very small number of cases that need prediction in this dataset (cf. Table \ref{tbl:annotStats}).
As regards the \textsc{STRICT} approach, we see that it highly improves precision from 65.7\% (of the top performing baseline) to 73\%, however with the cost of a high drop of recall (from 25\% to almost 10\%). 

These results demonstrate that the proposed \textit{MetaEL+} methods can significantly improve the performance of the individual systems and achieve results that are even competitive to
recent EL systems that make use of neural models, like \cite{cao2018neural} and \cite{kolitsas2018end} that report an F1 score of 80\% and 82.4\%, respectively, on the \textsc{CONLL} dataset.

\subsection{Prediction Performance}
\label{subsec:predPerformance}

\subsubsection{Multi-label classification} 
Table \ref{tbl:multilabelPerf} shows the prediction performance of our multi-label classifier.
We see that \textit{Jaccard Index} (ratio of correctly predicted labels) is high for CONLL (50.5\%) and IITB (58.7\%) but low for NEEL (36.5\%).
\textit{Hamming Loss} (ratio of wrong labels) ranges from 26.9\% (for IITB) to 41.3\% (for CONLL). With respect to the most strict metric \textit{Exact Match}, the score is 36.1\% for CONLL, 54.0\% for IITB, and 30\% for NEEL. In general, we see that the classification performance is very good for IITB and satisfactory for CONLL.
As we have already stressed (cf. Sect. \ref{subsubsec:evalMetricsClassPerf}), these metrics evaluate the correct prediction of all class labels per instance.
The \textit{real prediction accuracy} (last row of Table \ref{tbl:multilabelPerf}) shows the classification performance when considering if the correct entity is provided by the predicted system. 
The score is more than $90\%$ for CONLL and IITB, and almost 70\% for NEEL. These results demonstrate the high performance of our multi-label classifier. 

Looking now at the per-class performance, we see that for CONLL, the class label \textit{Babelfy} achieves the highest F1 score (69.8\%) while the \textit{TagMe} class has the lowest score (61.2\%). 
On the contrary, in IITB the \textit{Ambiverse} class achieves the highest F1 score (60.5\%), due to its very high precision (90.0\%), and \textit{TagMe} the lowest (45.1\%).
In NEEL, the highest F1 score is again achieved by the \textit{Ambiverse} class (45.0\%), however the lowest by the \textit{Babelfy} class (31.9\%).
These results show that there is no class for which the classifiers have a consistent high performance.

\begin{table}[]
\centering
\centering													
 \caption{Performance of multi-label classification}
\vspace{-4mm}	
\renewcommand{\arraystretch}{0.6}
\setlength{\tabcolsep}{4.5pt}
\small
\label{tbl:multilabelPerf}
\begin{tabular}{@{}lcccc@{}}
\toprule
 \textbf{Evaluation metric}  & \textbf{CONLL} & \textbf{IITB} & \textbf{NEEL} \\ \midrule

  Jaccard Index (\%) & 
  50.5 & 58.7  & 36.5  \\
  
  Hamming Loss (\%) & 
  41.3 &  26.9  & 29.9  \\ 
  
  Exact Match (\%) & 
  36.1 & 54.0  &	30.0  \\ \midrule
   
  Precision (\%) of Ambiverse class & 
  76.0 & 90.0  & 45.7  \\
  
  Precision (\%) of Babelfy class & 
  81.8 & 86.4  & 23.1 \\
  
  Precision (\%) of TagMe class &
  60.6 & 62.0 & 68.2 \\ \midrule
  
  Recall (\%) of Ambiverse class & 
  60.4 &  46.0 & 44.4   \\
  
  Recall (\%) of Babelfy class & 
  60.8 &  35.4 &    51.7  \\
  
  Recall (\%) of TagMe class &
  61.9 & 35.4 &  26.2 \\ \midrule
  
  F1 (\%) of Ambiverse class & 
  67.3 & 60.5 & 45.0  \\
  
  F1 (\%) of Babelfy class & 
  69.8 & 50.2 &  31.9  \\
  
  F1 (\%) of TagMe class &
  61.2 & 45.1 &  37.9 \\ \midrule
  \textbf{Real Prediction Accuracy (\%)} &  91.1 & 95.9  & 69.6 \\
  \bottomrule														
\end{tabular}	
\vspace{-5mm}	
\end{table}

\subsubsection{Binary classification} 
Table \ref{tbl:binClassPerf} shows the performance of the three binary classifiers used by the \textsc{MetaEL+$_{\text{STRICT}}$} method. First, we should highlight that the class distribution is very unbalanced. On average, around 78\% of the annotations are correct (\textit{true} class) and 22\% are wrong (\textit{false} class). This means that the \textit{false} class is underrepresented, which makes the classification problem harder. 

As expected, we notice that precision is very high for the majority \textit{true} class in almost all cases, while recall is high for the minority \textit{false} class. In CONLL, for example, precision of the \textit{true} class ranges from 88.6\% (TagMe classifier) to 91.5\% (Babelfy), while that of the \textit{false} class ranges from 25.5\% (Ambiverse) to 30.5\% (TagMe). On the contrary, recall of the \textit{true} class ranges from 45\% (Ambiverse classifier) to 54.2\% (TagMe) and of the \textit{false} class from 40.4\% (Babelfy) to 79\% (Ambiverse). 
Looking at the macro-averaged F1 scores, we notice that their performance is close to 50\% in almost all cases. 
TagMe classifier has the best performance in the two largest datasets (CONLL, IITB), however it has the worst performance in NEEL.
It is evident from these results that there is much room for further improvement for binary classification.

\begin{table}
\centering
\caption{Performance of binary classification.}		
\vspace{-4mm}
\renewcommand{\arraystretch}{0.6}
\setlength{\tabcolsep}{4.5pt}
\small
\label{tbl:binClassPerf}
\begin{tabular}{@{}lcccc@{}}
\toprule
 \textbf{Evaluation metric}  & \textbf{CONLL} & \textbf{IITB} & \textbf{NEEL} \\ \midrule
 
  Ambiverse - Precision (\%) of \textit{true} class     &  89.9 & 98.2 &  89.5  \\
  Ambiverse - Precision (\%) of \textit{false} class    &  25.5 & 21.4 &  31.0  \\ \midrule
  Ambiverse - Recall (\%) of \textit{true} class        &  45.0 & 46.6 &  45.9  \\
  Ambiverse - Recall (\%) of \textit{false} class       &  79.0 & 94.4 &  81.8 \\ \midrule
  Ambiverse - F1 (\%) of \textit{true} class            &  59.9 & 63.2 &  60.7  \\ 
  Ambiverse - F1 (\%) of \textit{false} class           &  38.6 & 34.9 &  45.0  \\ 
  \midrule \midrule
   
  Babelfy - Precision (\%) of \textit{true} class     &  91.5 & 96.5 &  93.8  \\
  Babelfy - Precision (\%) of \textit{false} class    &  27.2 & 66.3 &  51.7  \\ \midrule
  Babelfy - Recall (\%) of \textit{true} class        &  52.2 & 30.2 &  51.7 \\
  Babelfy - Recall (\%) of \textit{false} class       &  40.4 & 79.5 &  66.7  \\ \midrule
  Babelfy - F1 (\%) of \textit{true} class            &  66.5 & 46.0 &  66.7  \\ 
  Babelfy - F1 (\%) of \textit{false} class           &  38.6 & 34.9 &  45.0  \\ 
  \midrule \midrule
  
  TagMe - Precision (\%) of \textit{true} class     &  88.6 & 92.9 &  66.7 \\
  TagMe - Precision (\%) of \textit{false} class    &  30.5 & 35.5 &  32.3 \\ \midrule
  TagMe - Recall (\%) of \textit{true} class        &  54.2 & 33.2 &  20.9 \\
  TagMe - Recall (\%) of \textit{false} class       &  74.3 & 93.5 &  78.3 \\ \midrule
  TagMe - F1 (\%) of \textit{true} class            &  67.3 & 48.9 &  31.9  \\  
  TagMe - F1 (\%) of \textit{false} class           &  43.3 & 51.4 &  45.8  \\ 
  \midrule \midrule
   Ambiverse - Macro-averaged F1 (\%) & 49.3  & 49.1 & 52.9   \\
   Babelfy - Macro-averaged F1 (\%)   & 52.6  & 40.5 & 55.9   \\
   TagMe - Macro-averaged F1 (\%)     & 55.3  & 50.2 & 38.9   \\ 
  \bottomrule
\end{tabular}
\vspace{-4mm}
\end{table}

\subsection{Feature Analysis}

Table \ref{tbl:featureAnalysis} shows the EL performance for different combinations of features when considering the largest ground truth dataset (CONLL) and the best performing \textit{MetaEL+} method (\textsc{MetaEL+$_{\text{LOOSE}}$}).

With respect to the categories of features, we notice that the best performance is achieved when all categories are combined, which means that all contribute on achieving a high performance.
Regarding each individual category, we see that the \textit{surface form-based} features have the best performance, achieving an F1 score of 80.4\%. The \textit{mention-based} and \textit{document-based} features achieve 77.6\% and 78.2\%, respectively. 
The best pair of feature categories is the \textit{surface form-based} and \textit{document-based} (81\% F1) and the worst pair is the \textit{mention-based} and \textit{document-based} (77.6\% F1). These results show that the \textit{surface form-based} features have the highest contribution on achieving a good EL performance, and the \textit{mention-based} features the lowest contribution. 

Regarding the influence of each individual feature, we notice that $s_{ratio}$  (surface form's ratio of correct disambiguations per EL system) has the highest effect when we exclude it, dropping the F1 score from 81.9\% to 81\%. The second most influential feature is $m_{pos}$ dropping the F1 score to 81.1\%, which means that the mention's position in the document is a good indicator for the system that provides the correct entity.

\begin{table}[h]
\vspace{-3mm}
\small
\centering
\renewcommand{\arraystretch}{0.6}
\setlength{\tabcolsep}{4.5pt}
\caption{Effectiveness of different feature combination using \textsc{MetaEL+$_{{\text{LOOSE}}}$} on CONLL.}
\vspace{-4mm}
\label{tbl:featureAnalysis}
\footnotesize
\begin{tabular}{p{4.5cm}ccc}	
     \toprule
     \textbf{Features} & \textbf{P (\%)} & \textbf{R (\%)}    & \textbf{F1 (\%)} \\ 
     \midrule
All features	&	\textbf{84.8}	&	\textbf{79.2}	&	\textbf{81.9}	\\	\midrule
Only surface form-based	&	83.2	&	77.7	&	80.4	\\	
Only mention-based	&	80.3	&	75.1	&	77.6	\\	
Only document-based	&	80.9	&	75.6	&	78.2	\\	\midrule
Surface form-based + mention-based	&	83.5	&	78.0	&	80.7	\\	
Surface form-based + document-based	&	83.8	&	78.3	&	81.0	\\	
Mention-based + document-based	&	80.3	&	75.1	&	77.6	\\	\midrule
All features except $s_{words}$	&	84.4	&	78.9	&	81.5	\\	
All features except $s_{f}$	&	84.0	&	78.5	&	81.2	\\	
All features except $s_{df}$	&	84.5	&	78.9	&	81.6	\\	
All features except $s_{cand}$	&	84.4	&	78.9	&	81.5	\\	
All features except $s_{corr}$	&	84.3	&	78.8	&	81.5	\\	
All features except $s_{ratio}$	&	83.8	&	78.3	&	81.0	\\	
All features except $m_{pos}$	&	83.9	&	78.4	&	81.1	\\	
All features except $m_{sent}$	&	84.1	&	78.6	&	81.3	\\	
All features except $d_{words}$	&	84.5	&	78.9	&	81.6	\\	
All features except $d_{ents}$	&	84.0	&	78.5	&	81.2	\\	 	\bottomrule
\end{tabular}
\vspace{-6mm}
\end{table}

\subsection{Synopsis and Limitations}

The evaluation results can be summarised as follows: 
\begin{itemize}[noitemsep,nolistsep]
    \item Combining multiple EL tools through a \textit{MetaEL} approach can achieve a significantly better EL performance than individual systems in isolation.
    
    \item The proposed supervised ensemble approach (\textit{MetaEL+}) significantly outperforms the individual EL tools and six baseline methods in the largest and most reliable datasets.
    
    \item A \textsc{STRICT} \textit{MetaEL+} method which predicts if the entity provided by an EL system is correct can further improve precision without significantly affecting recall.
    
    \item The proposed multi-label classifier
    achieves a prediction accuracy of >90\% in the two largest datasets of our evaluation (CONLL and IITB), demonstrating its effectiveness. 
    
    \item The proposed binary classifiers achieve a relatively low accuracy (F1 score $\approx$ 50\%), showing that there is much room for improvement of the \textsc{STRICT} \textit{MetaEL+} method. 
    
    \item All three categories of features contribute to achieving the highest performance. With respect to the individual features, $s_{ratio}$ (surface form's ratio of correct disambiguations by each EL system) and $m_{pos}$ (mention's normalised position in the document) seem to be the most influential features. 
\end{itemize}

Limitations of our work are mainly concerned with (i) the limited performance of the binary classifiers in the \textsc{STRICT} approach, and (ii) the need of corpus-specific training data.

\section{Related Work}
\label{sec:rw}

The survey in \cite{shen2015entity} presents a thorough overview of the main approaches to EL, while more recent works (like \cite{cao2018neural}, \cite{kolitsas2018end} and \cite{fang2019joint}) exploit the idea of neural networks and deep learning. 
To the best of our knowledge, \cite{ruiz2015combining}, \cite{corcoglioniti2016microneel} and \cite{canale2018novel} are the only previous works that focus on the related (yet different) problem of \textit{MetaEL}, i.e., on how to combine the outputs of multiple EL tools for providing a unified set of entity annotations.

\cite{ruiz2015combining} proposes a weighted voting scheme inspired by the ROVER method \cite{fiscus1997post}. This method ranks the candidate entities by considering the performance of the systems on a so-called ranking corpus. Two of our baselines consider this method.
\cite{corcoglioniti2016microneel} focuses on microposts and resolve conflicts by majority vote or, in the event of a tie, by giving different priorities to the annotations produced by each annotator. Two of our baselines consider this approach. % (\textit{Majority+Random} and \textit{Majority+Best}). 
\cite{canale2018novel} describes a framework to combine the responses of multiple EL tools which relies on the joint training of two deep neural models. However, this work is not applicable in our MetaEL problem since it makes use of external knowledge (pre-trained word embeddings and entity abstracts) as well as entity type information (a type taxonomy from each extractor), as opposed to  our \textit{MetaEL} task which only considers plain lists of entity annotations. % returned by end-to-end EL tools.

With respect to the related problem of \textit{named-entity recognition} (NER), i.e., the detection of named entities in a given text and their classification in predefined categories like Person or Location, several works investigate how to combine the results of multiple NER methods \cite{dlugolinsky2014combining,corcoglioniti2016microneel,si2005boosting,plu2016enhancing}.
\cite{dlugolinsky2014combining} tackles the problem of concept extraction in \textit{microposts} and proposes machine learning methods that make use of features describing the microposts for combining the results of different NER tools. \cite{corcoglioniti2016microneel} also focuses on microposts and trains a multi-class SVM classifier. 
\cite{si2005boosting} focuses on \textit{bio-medicine} and proposes three methods for combining the results of various bio-medical NER systems: i) majority vote, ii) unstructured exponential model that considers the performance of the systems on training data, and iii) conditional random field that models the correlation between biomedical entities. We use the first two methods as baselines in our experiments. 
Finally, \cite{plu2016enhancing} unifies the outputs of three different named-entity extraction models (dictionary, POS tagger, NER) in a specific order and merges the overlapping mentions.

A related line of research on the NER problem combines multiple classifiers through ensemble learning \cite{wu2003stacked,florian2003named,saha2013combining,speck2014ensemble}. 
\cite{wu2003stacked} examined several stacking and voting (majority-based) methods that combine three different classifiers. 
In a similar way, \cite{florian2003named} combines the results of four classifiers,
while \cite{saha2013combining} constructs an ensemble of seven classifiers. 
\cite{speck2014ensemble} evaluates the performance of 15 classification models, finding that ensemble learning can highly reduce the error rate of state-of-the-art NER systems. 
These works use as features the \textit{predictions} of multiple supervised classifiers for deciding on the entity \textit{type} of a given mention (from a pre-defined list of entity types), as opposed to our MetaEL task which combines \textit{EL systems} and considers features extracted from the underlying \textit{corpus} for training dedicated classifiers able to predict the \textit{EL system} that can provide the correct link for a given mention. 

A related interesting work is the NERD framework \cite{rizzo2011nerd} which allows running multiple EL systems on the same text(s). NERD uses a common ontology for storing the results, thus providing a common representation format and facilitating the evaluation of NER and EL methods. However, it does not resolve conflicts like in the case of MetaEL. Our work can be used by this framework for conflict resolution and for providing a single set of entity annotations.

\section{Conclusions and Future Work}
\label{sec:conclusion}

We have argued that the performance of entity linking (EL) on a given corpus may be optimised by combining the results of distinct EL tools. To this end, we introduced a novel approach towards \textit{Meta Entity Linking (MetaEL)} where outputs of multiple end-to-end EL tools are unified on a per-mention basis through an ensemble learning approach. We model the problem as a supervised classification task and provide a rich set of features that can be used within a supervised classifier for predicting the EL system that can provide the correct entity link for a given mention. 

Using existing ground truth datasets and three EL tools, we compared the performance of the proposed models with each individual EL tool and with six baseline methods. The results show that, considering the largest ground truth dataset (CONLL), our multi-label classifier significantly outperforms the F1 score of both the best performing individual EL system (by 10\%) and the best baseline (by 5\%). Using binary classification for cases where a mention is recognised by only one EL system, a more selective (\textsc{STRICT}) approach that predicts the correctness of the provided entity link can further improve precision without significantly affecting recall. 
Results on the performance of the prediction tasks per se demonstrated the effectiveness of the proposed multi-label classifier. Finally, an extensive feature analysis showed that all the proposed features contribute on achieving a high EL performance.

Given the promising results of our experiments, in the future we plan to extensively evaluate the performance gain of MetaEL using different number and combinations of EL tools, including more recent tools that make use of neural models. This will provide a better understanding of the circumstances under which MetaEL has a significant effect in the EL performance. 
We also intend to study distantly supervised approaches where weakly labelled training data are automatically generated based on heuristics, aiming at solving the problem of obtaining corpus-specific training data. Finally, we plan to investigate the applicability of more advanced models for the binary classification task, in order to improve its (relatively low) performance. 

\begin{acks}
This work was partially supported by CNPq (Brazilian National Council for Scientific and Technological Development) under grant GDE No. 203268/2014-8.
\end{acks}

\bibliographystyle{ACM-Reference-Format}
\bibliography{main}

\end{document}